\pdfoutput=1
\documentclass[11pt]{article}

\usepackage{ACL2023}

\usepackage{times}
\usepackage{latexsym}

\usepackage[T1]{fontenc}
\usepackage[utf8]{inputenc}

\usepackage{microtype}

\usepackage{inconsolata}

\usepackage{graphicx}
\usepackage{multicol}
\usepackage{multirow}
\usepackage{booktabs}
\usepackage{bbding}
\usepackage{amssymb, amsmath}
\usepackage{threeparttable}
\def\eg{\textit{e.g.}}

\title{AUGUST: an Automatic Generation Understudy\\ for Synthesizing  Conversational Recommendation Datasets}
\author{Yu Lu$^{2,1}$, \hspace{0.1cm}
Junwei Bao$^{3}\thanks{~~Corresponding author: baojunwei001@gmail.com}$, \hspace{0.1cm}
Zichen Ma$^{2,1}$, \hspace{0.1cm}
Xiaoguang Han$^{1,2}$,\\
\bf Youzheng Wu$^{3}$, \hspace{0.1cm}
Shuguang Cui$^{1,2}$, \hspace{0.1cm}
Xiaodong He$^{3}$\\
  $^1$ SSE, CUHKSZ \quad $^2$ FNii, CUHKSZ \quad $^3$JD AI Research\hspace{0.8cm}  \\
  $^{2,1}$\{yulu1,zichenma1\}@link.cuhk.edu.cn\\
  $^{3}$\{baojunwei,wuyouzheng1,xiaodong.he\}@jd.com\\
  $^{1,2}$\{hanxiaoguang,shuguangcui\}@cuhk.edu.cn\\
  }

\begin{document}
\maketitle
\begin{abstract}
High-quality data is essential for conversational recommendation systems and serves as the cornerstone of the network architecture development and training strategy design. Existing works contribute heavy human efforts to manually labeling or designing and extending recommender dialogue templates. However, they suffer from (i) the limited number of human annotators results in that datasets can hardly capture rich and large-scale cases in the real world, (ii) the limited experience and knowledge of annotators account for the uninformative corpus and inappropriate recommendations. In this paper, we propose a novel automatic dataset synthesis approach that can generate both large-scale and high-quality recommendation dialogues through a data2text generation process, where unstructured recommendation conversations are generated from structured graphs based on user-item information from the real world. In doing so, we comprehensively exploit: (i) rich personalized user profiles from traditional recommendation datasets, (ii) rich external knowledge from knowledge graphs, and (iii) the conversation ability contained in human-to-human conversational recommendation datasets. Extensive experiments validate the benefit brought by the automatically synthesized data under low-resource scenarios and demonstrate the promising potential to facilitate the development of a more effective conversational recommendation system\footnote{Our code will be released in \url{https://github.com/JD-AI-Research-NLP/AUGUST}}.
\end{abstract}

\section{Introduction}
\label{intro}

Conversational recommendation (CR) systems aim to recommend potential items of interest for users (or seekers) through dialogue-based interactions. Although tremendous works have been contributed to the CR domain, the lack of both large-scale and high-quality training data remains a common problem due to the great cost and difficulty in dataset construction. A classic recommendation dialogue collection \cite{li2018towards} relies on a human recommender to chat with a randomly paired seeker and supply some recommendations within several conversation turns usually based on the chatting content. The dataset constructed under this paradigm is not only limited in scale but also can hardly ensure the recommendation quality. Specifically, it suffers from: (i) the limited number of human annotators that  results in datasets can hardly capture rich and large-scale cases in the real world, (ii) the limited experience and knowledge of annotators account for the uninformative corpus and inappropriate recommendations. In addition, the preference given by annotators to the recommended item may be ``unreal'' when (s)he is unfamiliar with it but cannot timely validate the annotation. The performance of a CRS trained with such datasets may be barely satisfactory when applied in real-world scenarios. % Therefore, such artificially generated CR data may not enable the neural learning models to distill sufficient knowledge of recommendation and related conversation ability for application in real-world scenarios. 

\begin{figure}[t]%H为当前位置，!htb为忽略美学标准，htbp为浮动图形
\centering %图片居中
\includegraphics[width=0.47\textwidth]{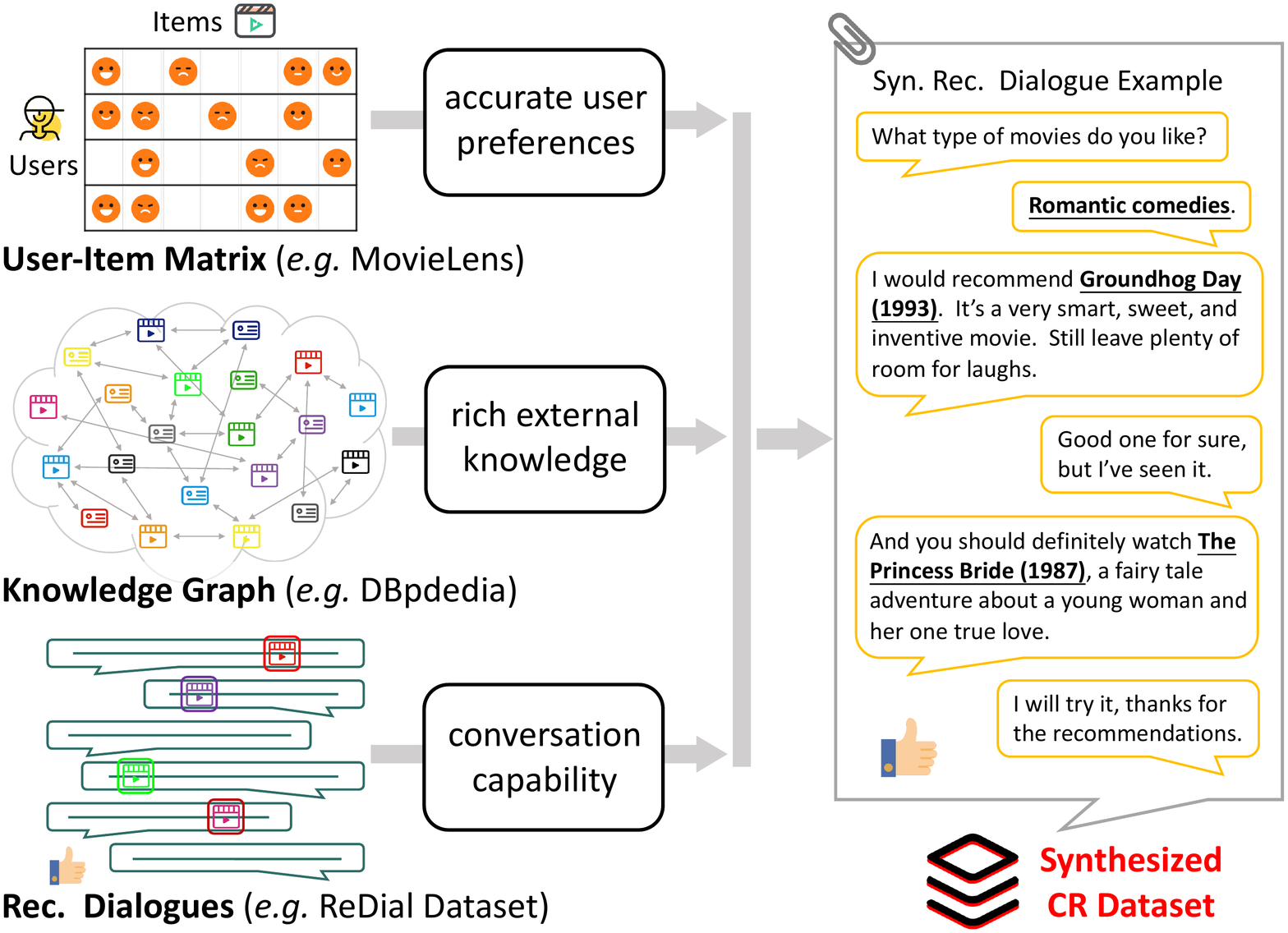}%插入图片，[]中设置图片大小，{}中是图片文件名
\caption{The proposed approach takes three kinds of sources, namely user-item matrices, knowledge graphs, and existing conversational recommendation datasets, to automatically generate recommendational dialogues.} %最终文档中希望显示的图片标题
\label{Fig.main1} %用于文内引用的标签
\end{figure}

Although there exist numerous recommendation data that contain more ``real-world'' user preferences, \eg, MovieLens~\cite{harper2015movielens}, there are little or even no corresponding dialogues, which leads to a low-resource scenario for CRS training. Therefore, we propose a novel CR data synthesis approach, AUGUST, which is an \underline{AU}tomatic \underline{G}eneration \underline{U}nder\underline{ST}udy for conversational recommendation datasets. The core of our approach is synthesizing the strengths from three kinds of data resources: (i) user-item ratings from websites that can provide items really favored by each user; (ii) external knowledge that can provide rich item-related information leading to a ``professional'' recommender; (iii) abundant dialogue corpus that can help develop the learning model's conversation ability. Note that all three can be easily accessed thus facilitating the potential of generating large-scale diverse recommendational dialogues. In doing so, our approach contains two steps: (1) to form one data sample, seamlessly selecting some items rated by one user, from which a graph is constructed that contains the items, related entities, and their relations based on a well-developed knowledge graph (KG); (2) adopting a Data2Text generator~\cite{li2021few} to convert the item graph into a fluent and natural dialogue around the items. Such a graph-based dialogue generation manner is endowed with great extensibility and explainability where external knowledge can be integrated via expanding the intermediate graph with related entities from KG. To train the Data2Text module, we make use of recommendational dialogues from existing CR datasets to learn a dialogue generator. Specifically, we elicit graphs from dialogues as ones from user-item ratings, and train the Data2Text generator to take the graph as input to recover the original dialogue. 
% current well pre-trained language model and finetune it on existing CR datasets by taking extracted items to recover the original dialogue. 

We conduct extensive experiments on the synthesized data quality and the performance of Data2Text generation, and give a detailed analysis of problems in the synthesis process. We also empirically validate the benefit of synthesized data in helping learn a stronger CRS, especially on recommendation accuracy in the low-resource scenario. Along with the rapid development of Data2Text generation methods, the proposed AUGUST is of great potential and provides a new solution to construct large-scale CR datasets, which is our main contribution. In addition, it is expected to attract more attention to the direction of automatic dataset generation, and facilitate the data-driven learning models designed for not only CR but also other various tasks in the future. 
\section{Related Work}
%组织形式1： CR(data+models) + D2T(data+models) 
%该形式下D2T 展开来讲意义不大，主要应该聚焦于D2T的发展现状
%组织形式2： CRD + D2T + DA
%该形式下可以说明 现有dataset的问题 + D2T的发展进程 + 引出利用redial数据训movielens需要借助DA的method和理论（增加一些technic)
% \subsection{Recommendation Systems}
% Recommendation systems aims at accurately and timely helping users to find items of interest. Traditional recommendation systems primarily predict a user's preference by analyzing past behaviors offline, \eg, click history, visit log, and ratings on items. Early methods, such as collaborative filtering \citep{} and factorization machine \citep{} have been intensively used in practical applications due to its efficiency and interpretability.  Conversational recommendation systems take a different approach and support a richer set of interactions. These interactions can, for example, help to improve the preference elicitation process or allow the user to ask questions about the recommendations and to give feedback. With the rapid increasing interest in CRS, more datasets are proposed to facilitate the research of method in this region. Some studies \citep{} collect human-human and human-machine conversation data by asking true users to converse using natural language under certain rules.
% \citet{zhou2021crslab} have implemented an open-source toolkit, called CRSLab, for building and evaluating CRSs, \citet{gao2021advances} presented that the sacle of existing datasets is not enough to develop the CRSs
% \paragraph{Traditional Recommendation Systems}
\subsection{Conversational Recommendation Dataset}
Recently, Conversational Recommendation Systems (CRS) \cite{li2018towards,chen2019towards,jannach2021survey,lu2021revcore} have become an emerging research topic, which aims to provide high-quality recommendations to users through natural language. To facilitate the study of this task, some works collect human-human and human-machine conversation data by asking human annotators to conversate under certain rules. \citeauthor{hayati2020inspired} manually annotate each utterance with the sociable strategies to validate the effectiveness of sociable recommendation strategies in CRS. \citeauthor{moon2019opendialkg} present a parallel dialog$\leftrightarrow$KG corpus where each mention of an entity is manually linked with its corresponding KG paths. \citeauthor{liu2020towards}  create a multi-type dialogue dataset and want the bots can proactively and naturally lead a conversation from a non-recommendation dialogue to a recommendation dialog. Similarly, \citeauthor{zhou2020towards} proposes a topic-guided CR dataset to help the research of topic transitions.   However, \citeauthor{gao2021advances} point that existing datasets are not qualified to develop CRS that satisfies industrial application requirements for two reasons: 1) the scale of these datasets is not enough to cover the real-world entities and concepts; 2) the datasets constructed under certain rigorous constraints can hardly generalize to the complex and diverse real-world conversation. Therefore, more efforts are encouraged to develop large-scale, generalizable, and natural datasets for CRS. 
% \paragraph{user preference elicitation}
% \paragraph{Conversational Recommendation Systems}
\subsection{Data2Text Generation}
%Data-To-Text Generation is a task of automatically producing text from non-linguistic input.
Data2Text Natural Language Generation (NLG) is the computational process of generating meaningful and coherent natural language text to describe non-linguistic input data. %Practical applications can be found in domains such as weather forecasts~\cite{}, health care~\cite{}, feedback for drivers~\citep{}, diet management~\cite{}, election results~\cite{} and sportscasting news~\cite{}.
The input can be in various forms such as databases of records, spreadsheets, knowledge bases, and simulations of physical systems. Traditional methods for Data2Text generation~\cite{reiter2000building} implement a pipeline of modules including content planning, sentence planning, and surface realization. With the rapid development of Seq2Seq models especially pre-trained models, recent neural generation systems~\cite{li2021few} trained in an end-to-end fashion get state-of-the-art results on Data2Text benchmarks such as WebNLG~\cite{gardent2017webnlg}, ToTTo~\cite{parikh2020totto}, and AGENDA~\cite{koncel2019text}. %There is some work evaluating and analyzing the data-to-text generation task.  
One of the most popular subtasks, Graph2Text, aims to create fluent natural language text to describe an input graph. Early works mainly center around statistical methods, applying grammar rule to generate text~\cite{konstas2013inducing}. Recently, neural-network-based approaches have been proposed to generate text from linearized KG triples~\cite{ferreira2019neural}, some of which investigate how to encode the graph structural information using Graph Neural Networks (GNNs)~\cite{scarselli2008graph} and Transformer~\cite{koncel2019text} explicitly. Unsupervised methods~\cite{guo2020cyclegt} and few-shot problems~\cite{li2021few} are also explored. In our approach, we adopt a Graph2Text generator for CR data synthesis. 

% \paragraph{AMR-to-text generation}  AMR-to-text generation  
%\subsection{Knowledge-enhanced Generation}

\begin{figure*}[htbp]%H为当前位置，!htb为忽略美学标准，htbp为浮动图形
\centering %图片居中
\includegraphics[width=0.85\textwidth]{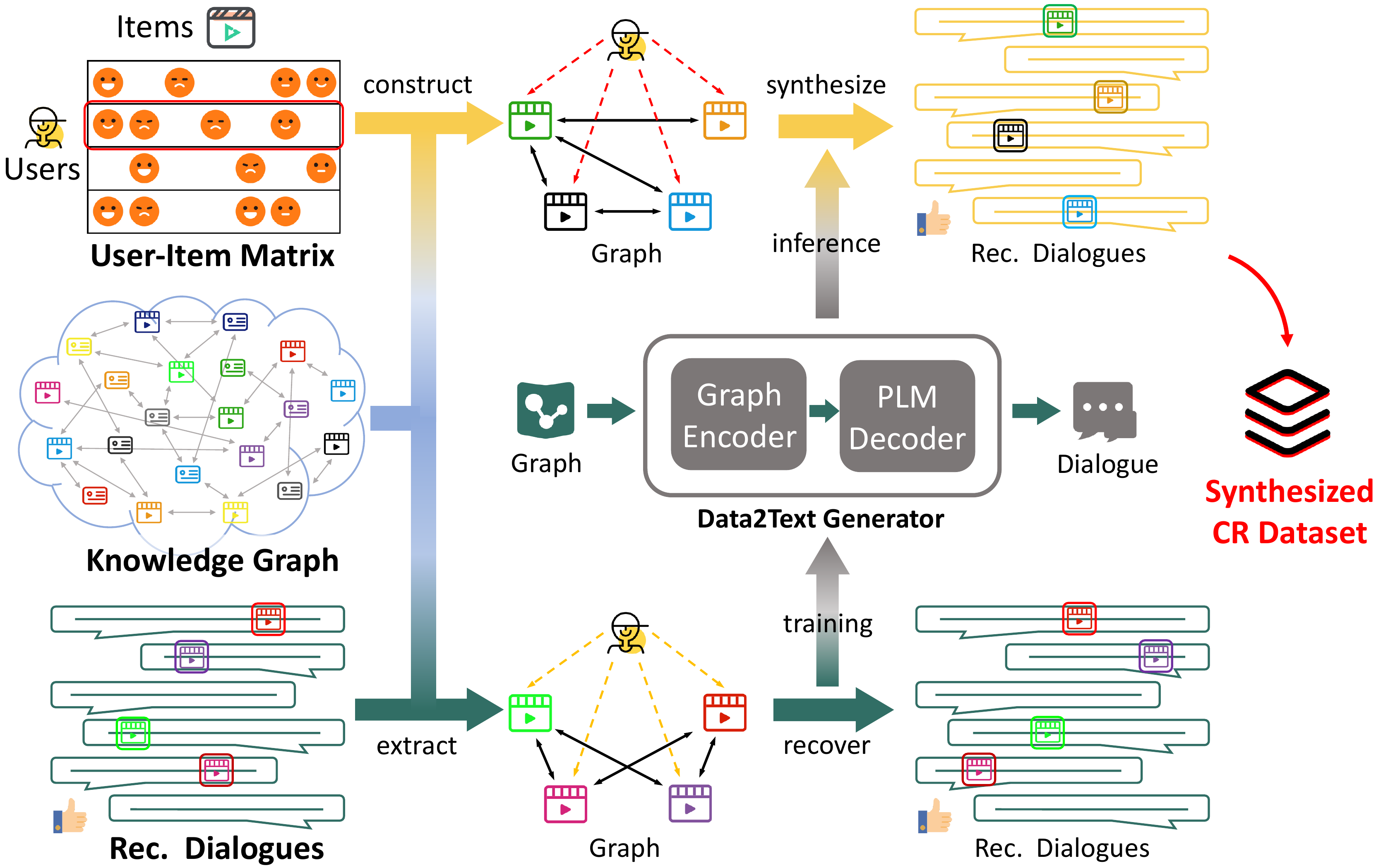}%插入图片，[]中设置图片大小，{}中是图片文件名
\caption{The overview of the proposed AUGUST framework for automatic recommendational dialogue synthesis. } %最终文档中希望显示的图片标题
\label{Fig.method} %用于文内引用的标签
\end{figure*}
\section{Methodology}
%介绍训练集和测试集构造方法，以及构造过程中的策略
%介绍我们提出的一种evaluation metric作为辅助
\subsection{Preliminaries}
Our CR dataset synthesis approach produces recommendational dialogues from three kinds of resources: user-item matrices from traditional recommendation datasets, external knowledge graphs, and existing CR datasets. We first introduce related notations. \textbf{A user-item matrix} (UIM) $\mathbf{M}$ (supplied by datasets like MoviLens~\cite{harper2015movielens}) consists of $N$ rows and $M$ columns, of which the $i$-th row represents the ratings of the $i$-th user $\mathrm{U}_i$ towards all $M$ items, and each element $s_{ij} \in [1, 2, 3, 4, 5]$ represents the $i$-th user's rating score towards the $j$-th item $o_j$, where a higher score represents the user's more favor to one item. Note that the matrix $\mathbf{M}$ may be sparse depending on the number of ratings given by each user. 
\textbf{A knowledge graph} $\mathbf{G} = <\mathcal{E}, \mathcal{R}>$, \eg~DBpedia~\cite{auer2007dbpedia}, where $\mathcal{E}$ and $\mathcal{R}$ are the entity and relation set, respectively. The graph consists of large amounts of entity-relation-entity triples $(e_i, r_{ij}, e_j)$, of which $e_{i}$ or $e_{j}$ can be an item or non-item entity from $\mathcal{E}$ and $r_{ij} \in \mathcal{R}$ represents the relation category between an associated entity pair. We denote the item entity set as $\mathcal{O} \subset \mathcal{E}$, which contains all recommendation candidates. 
In \textbf{a CR dataset}, \eg~the ReDial dataset~\cite{liu2020towards}, a conversation is generated for recommendations on a certain domain (movie, traveling, or restaurant, etc.) in a seek-recommender pair. Denote the $i$-th conversation as $\mathrm{C}_i$, a seeker/user $\mathrm{U}_i$ is asking for item recommendations from a recommender $\mathrm{R}_i$. In the following chatting turns, $\mathrm{U}_i$ may express his/her preferences explicitly or implicitly, then $\mathrm{R}_i$ is expected to capture the user's preferences according to the historical dialogue context, denoted as $\mathcal{C}_t=\{c_j\}_{1}^t$, where $t$ is the historical turn number and $c_j$ is the $j$-th conversation utterance. %At the (t−1)(t-1)-th turn, one should produce the next utterance ctc_{t} to reply the other. The reply of Ri\mathrm{R}_i can cover one or several candidate items {oj}\{o_j\} selected from the entire item set O\mathcal{O} or ask other attributes of items according to the strategy. %note that Ot=O1t,…,OTt\mathcal{O}_t = {O_t^1,\dots, O_t^T} can be equal to \empty\empty when there is no need to recommendation. In such a case, then hih_i may raise a clarification question or generate a chit-chat response.   

\subsection{Dataset Synthesis}
\label{synthesis}
The proposed dataset synthesis approach starts from real-world user preferences information easily accessed from the UIM $\mathbf{M}$. Then a UIM$\rightarrow$Graph$\rightarrow$Dialogue generation pipeline is adopted to synthesize recommendational dialogues, with the overview shown in Fig.~\ref{Fig.method}. 

\paragraph{UIM $\rightarrow$ Graph}
The first step is to convert UIM that contains user preferences into graphs. From any row $i$ of $\mathbf{M}$, a set of items with respective ratings $\{(o_j, s_{ij})\}$ can be taken to generate a dialogue sample. All $o_j$ are used as nodes to construct the graph $\mathbf{G}'_i$. To integrate the user preferences into $\mathbf{G}'_i$, an extra node of user $u_i$ with its relation to each item node is added to constitute triples like $(u_i, s_{ij}, o_j)$ for item $o_j$. Furthermore, we extend $\mathbf{G}'_i$ by incorporating rich external knowledge from $\mathbf{G}$ for the informativeness of the final dialogue output. Specifically, for each two items $o_j$ and $o_k$, we search for a two-hop path in $\mathbf{G}$ to find their relations, \textit{i.e.}, two movies are directly linked (neighbouring) as $(o_j, r_{jk}, o_k)$ (\eg~belong to one movie series) or linked by one entity $e_l$ as $(o_j, r_{jl}, e_l, r_{lk}, o_k)$ (\eg~sharing the same director, actors, or genre). Then, these triples in the searched paths are added into $\mathbf{G}'_i$. The obtained graph $\mathbf{G}'_i$ can better represent the selected items from UIM data by incorporating both accurate user-preference information and knowledge-equipped inter-entity relations. 

\paragraph{Graph $\rightarrow$ Dialogue}
Given a graph $\mathbf{G}'_i$ that represents the items expected to appear in the dialogue, a Data2Text generator aims to synthesize a conversational dialogue $\mathrm{C}_i$ based on the graph. We cast it as a Data2Text problem. We adopt a Data2Text generator to take the graph as input, and output raw text that contains the vertex and edge information in the graph. Note that two tokens [U] (user) and [R] (responder) are specially defined to be generated in the text, such that the sentences after [U] ([R]) and before the next token [R] ([U]) can be viewed as a single turn. In this way, the text can be decomposed and re-organized into a multi-turn dialogue. Considering there is no supervision (graph-dialogue data pair) for the learning of the generator in this Data2Text process, we utilize the conversation corpus in existing CR datasets to learn a strong generator for dialogue synthesis, which is introduced in the following subsection.

\subsection{Data2Text Generation}
\label{data2textDA}
\begin{figure}[t]%H为当前位置，!htb为忽略美学标准，htbp为浮动图形
\centering %图片居中
\includegraphics[width=0.47\textwidth]{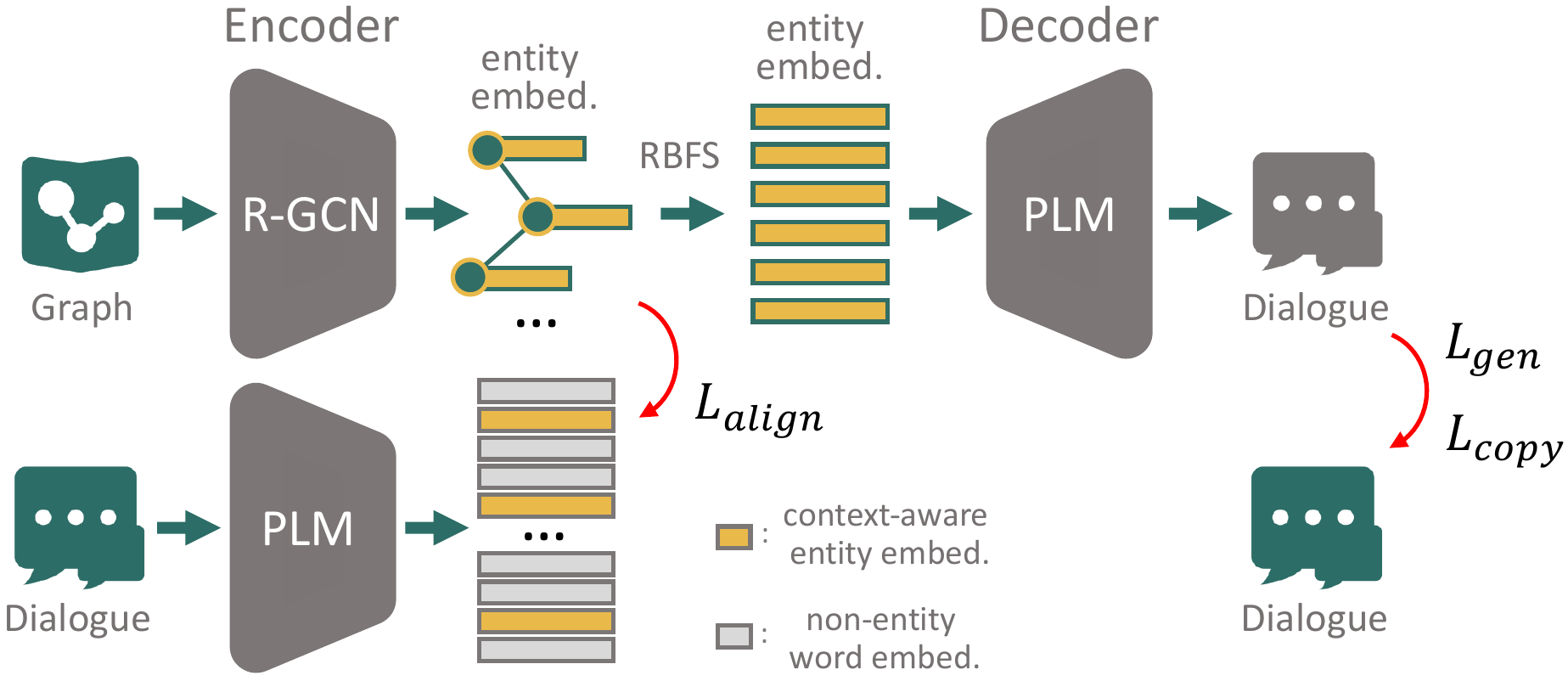}%插入图片，[]中设置图片大小，{}中是图片文件名
\caption{The illustration of the used encoder-decoder architecture for Data2Text generation.} %最终文档中希望显示的图片标题
\label{Fig.autoencoder} %用于文内引用的标签
\end{figure}

In order to generate both natural and logical dialogues from item-related graphs, we adopt a Data2Text generator to learn the conversation knowledge in existing CR datasets for Graph $\rightarrow$ Dialogue generation. As illustrated in Fig.~\ref{Fig.autoencoder}, an encoder-decoder architecture is implemented with an R-GCN encoder~\cite{schlichtkrull2018modeling} for graph feature extraction, and a pre-trained language model (PLM)~\cite{lewis2020bart} decoder for dialogue generation.
\paragraph{Graph Construction and Encoding}
Given any dialogue sample $\mathrm{C}_i$ in existing CR datasets, we construct a graph $\mathbf{G}'_i$ to produce a graph-dialogue training pair for learning a strong Data2Text generator. To construct $\mathbf{G}'_i$ from $\mathrm{C}_i$, we first search for all entities $\{e_j\}$ with the speaker's ($\mathrm{U}_i$ or $\mathrm{R}_i$) sentiment $\{s_{ij}\}$ to them (provided by CR datasets usually or generated from an estimator), and link each $e_j$ with corresponding nodes in $\mathbf{G}$. Then a graph $\mathbf{G}'_i$ can be constructed in a similar way as in the UIM $\rightarrow$ Graph process described in Sec.~\ref{synthesis}. %In the next step, we adopt an encoder-decoder architecture for Data2Text generation. 
Given a constructed $\mathbf{G}'_i$, an R-GCN~\cite{schlichtkrull2018modeling} is applied as the encoder to generate entity embeddings for $\mathbf{G}'_i$. Let $\phi_j \in \mathbb{R}^d$ denote the entity embedding for a general entity $e_j$ in KG, where $d$ is the embedding size. Then the R-GCN helps leverage the multi-relational information to have a structure-aware graph representation. Specifically, the embedding of $e_j$ at the $l+1$-th of total $L$ layers can be computed as:
$$
\phi_j^{l+1} = \sigma(\sum_{r\in\mathcal{R}}\sum_{k\in\mathcal{N}_j^r} \mathbf{W}_r^l\phi_k^l + \mathbf{W}_0^l\phi_j^l),
$$
where $\sigma(\cdot)$ is the activation function,  $\mathbf{W}_r^l$ and $\mathbf{W}_0^l$ are trainable parameters, and $\mathcal{N}_j^r$ is the set of neighbouring entities of $e_j$ under relation $r$. Note that, all $\phi_j^0$ before the first layer are initialized by pre-trained KG embeddings in \cite{yang2014embedding}. The entity embeddings $\{\phi_j^L\}$ output by the last R-GCN layer are re-denoted as $\{\phi_j\}$ for simplification.

\paragraph{Graph Feature Learning} To learn higher-quality graph features for more smooth decoding, we leverage another encoding branch of a pre-trained language model (PLM) to learn context-aware node features and align ones encoded from graphs with them. Specifically, by taking the whole dialogue as PLM input, entities are represented with contextual information in natural utterances, so that rich knowledge in PLM can be adapted. Denote the context-aware entity embedding output by the PLM branch as $\hat{\phi_j} \in \mathbb{R}^d$, which has the same dimension as the R-GCN embedding. The alignment between two types of entity feature vectors is implemented by minimizing an $l_2$ loss, denoted as $L_{align}$:
$$
L_{align} = \sum_{e_j \in \mathbf{G}'_i}||{\phi_j} - 
\hat{\phi_j} ||^2.
$$
Before feeding graph node features into the decoder, we linearize them into an entity sequence $\{\phi_j\}$ through a relation-biased breadth-first search (RBFS) strategy following \cite{li2021few}, where a breadth-first search is adapted and an RBFS weight $\alpha_j$ is computed for each node $e_j$ as its score to decide the order in each search level:
$$
\alpha_j = \sigma(\phi_i^\top ~\mathbf{W}_r^L~ \phi_j), <e_i, r, e_j> \in \mathbf{G}',
$$
where $e_i$ is the parent node of $e_j$ in the search process. In the same search level, the node with a higher RBFS score has a higher order in the sequence. For more related implementation details, please refer to \cite{li2021few}. 

\paragraph{Dialogue Decoding} In the decoding stage, a PLM decoder is performed to decode the linearized graph features $\{\phi_j\}$ into textual dialogues. To formalize the dialogue generation into a typical natural language generation problem, we sequentially connect all utterances into a single paragraph but with special tokens as the separation for regrouping into dialogue turns.  Denote the $k$-th of total $K$ tokens as $w_k$, the generation objective is to minimize the negative log-likelihood as:
$$
L_{gen} = -\sum_{k=1}^K \log P(w_k | w_1, w_2, \cdots, w_{k-1}), 
$$
where $P(\cdot)$ denotes the probability function. To encourage covering entities from the input graph, a copy mechanism implemented with a pointer network is conducted, leading to a copy loss term $L_{copy}$. 

The overall objective function to learn the domain adaptive encoder-decoder can be written as:
$$
L_{over} = L_{gen} + \lambda_1 L_{align} + \lambda_2 L_{copy},
$$
where $\lambda_1$ and $\lambda_2$ are weight factors to balance different loss terms, respectively.

% \paragraph{Adversarial Domain Adaptation}

\begin{table*}[]
\setlength{\tabcolsep}{2.35mm}{
\begin{tabular}{l|cccccc|cccc}
\toprule
% Test Data & Recall & B-2 & B-4 & R-L & CIDEr & Chrf & Dist-1 & Dist-2 & Dist-3 & PPL  \\  
\multirow{2}{*}{\textbf{Test Data}} & \multicolumn{6}{c|}{Reconstruction} &\multicolumn{4}{c}{Writing Quality}\cr
\cmidrule(lr){2-7} \cmidrule(lr){8-11}
& \textbf{Recall} & \textbf{B-2} & \textbf{B-4} & \textbf{R-L} & \textbf{CIDEr} & \textbf{Chrf} & \textbf{Dist-1} & \textbf{Dist-2} & \textbf{Dist-3} & \textbf{PPL}  \\
\midrule 
WebNLG  & - & 35.01 & 19.82 & 48.02 & 1.65 & 43.65 & 0.90 & 0.92  & 0.87 & 9.16   \\   
ReDial  & 0.82 & 28.44 & 11.43 & 28.82 & 1.52 & 42.16 & 0.43 & 0.75   & 0.84 & 3.39   \\   
ML-G2D  & 0.78 & 21.51 &  7.53 & 24.82 & 1.04 & 32.60 & 0.44 & 0.84   & 0.98 & 3.57   \\
\bottomrule          
\end{tabular}}
\caption{Performance of Data2Text generation on three datasets. B-$n$ denotes BLEU-$n$ and R-L denotes ROUGE-L. }
\label{tab:auto_eval}
\end{table*}
\begin{table}[]
\setlength{\tabcolsep}{1.mm}{
\begin{tabular}{l|cc|ccc}
\toprule
\multirow{2}{*}{\textbf{Data}} & \multicolumn{2}{c|}{Distinct.}  & \multicolumn{3}{c}{Language Nat.}    \\ 
\cmidrule(lr){2-3} \cmidrule(lr){4-6}
& \small \textbf{~~Dist-2~} & \small \textbf{~Dist-3~~} & \small \textbf{~~Logic~} & \small \textbf{Fluency}  & \small \textbf{Inform.} \\ \midrule
\small{AUGUST} & 2.7 & 4.2 & ~4.4 & 4.0 & 3.9\\
ReDial         & 2.8 & 4.4 & ~4.6 & 4.6 & 4.0\\
\bottomrule
\end{tabular}}
\caption{Comparison on Distinctness and Language Naturalness (via human evaluation) of AUGUST synthesized data and ReDial data. ``Inform.'' means informativeness.}
\label{eva:human}
\end{table}

\section{Experiments}
% In this section, we first set up the experiments, and then report the results and analysis.
\subsection{Experiment Setting}
\subsubsection{Resources}
% MovieLens, ReDial, Dbpedia
\textbf{(1) The ReDial dataset}~\cite{li2018towards} is collected by crowd-sourcing users on Amazon Mechanical Turk (AMT). Two paired workers serve as the recommender and user to produce a conversation and cover at least 4 different movies. Every movie mentioned in the dialog is annotated explicitly. ReDial contains 10,021 conversations related to 64,362 movies and is split into training, validation, and test sets with a ratio of 8:1:1. 
\textbf{(2) The MovieLens} dataset~\cite{harper2015movielens}, released by GroupLens Research, describes people's expressed preferences for movies. These preferences take the form of $<$user, item, rating, time-stamp$>$ tuples, where the rating (1$\sim$5) represents the user's preference for a movie at a particular time. These preferences are collected by the MovieLens website, a recommender system that asks its users to give movie ratings for personalized movie recommendations. 
\textbf{(3) The DBpedia knowledge base}~\cite{auer2007dbpedia} contains structured knowledge extracted from Wikipedia. It collects rich movie-related information and inter-movie relations and releases an open knowledge graph available to the public. % A knowledge graph is a special kind of database which stores knowledge in a machine-readable form and provides a means for information to be collected, organised, shared, searched, and utilised. 

\begin{table*}[!tp]
  \centering
%   \fontsize{6.5}{8}\selectfont
  \begin{threeparttable}
    \setlength{\tabcolsep}{2.35mm}
    \setlength{\abovecaptionskip}{2mm}
    %\begin{tabular}{p{14mm} p{5mm} p{5mm} p{5mm} p{8mm} p{8mm} p{8mm}}
    \begin{tabular}{c|c|c|ccc|cccc}
    \toprule
    \multirow{2}{*}{~~\textbf{ReDial}~~} & \multirow{2}{*}{\textbf{AUGUST}} & \multirow{2}{*}{~~\textbf{ML}~~} 
    &\multicolumn{3}{c|}{Recommendation} &\multicolumn{4}{c}{Conversation}\cr
    \cmidrule(lr){4-6} \cmidrule(lr){7-10}
    & & & \textbf{R@1} & \textbf{R@10} & \textbf{R@50} & ~~\textbf{Dist-2} & ~~\textbf{Dist-3} &  ~~\textbf{Dist-4} & \textbf{PPL} \\
    \midrule
    & &   0\% & 0.00 & 0.01 & 0.02 & ~~- & ~~- & ~~-& -     \\
    & &  50\% & 0.08 & 1.76 & 2.21 & ~~- & ~~- & ~~- & ~~-  \\
    & & 100\% & 0.17 & 1.59 & 2.94 & ~~- & ~~- & ~~- & ~~- \\
    \midrule
    \checkmark & &   0\%  & 0.00 & 1.77 & 3.53 & ~~0.292 & ~~0.336 & ~~0.470 & 14.2 \\ 
    \checkmark & &  50\%  & 0.00 & 2.65 & 6.19 & ~~0.303& ~~0.411 & ~~0.482 & 15.7 \\ 
    \checkmark & & 100\%  & 0.01 & 2.77 & 6.02 & ~~0.321 & ~~0.374 & ~~0.510 & 16.9 \\ 
    \midrule
    & \checkmark &   0\%  & 1.32 & 4.42 & 15.93 & ~~0.239 & ~~0.315 & ~~0.318 & \textbf{11.4} \\ 
    & \checkmark &  50\%  & \textbf{1.76} & 4.42 & 14.16 & ~~0.307 & ~~0. 316& ~~0.425 & 14.6\\ 
    & \checkmark & 100\%  & 0.88 &\textbf{7.79} & \textbf{19.46} & ~~0.297 & ~~0.301 & ~~0.412 & 13.8  \\ 
    \midrule
    \checkmark & \checkmark &   0\%  & 0.17 & 1.77 & 8.65 & ~~0.292& ~~0.375 & ~~0.451& 14.0 \\ 
    \checkmark & \checkmark &  50\%  & 0.84 & 2.66 & 10.31 & ~~\textbf{0.360} & ~~\textbf{0.451} & ~~0.507 & 15.7  \\ 
    \checkmark & \checkmark & 100\%  & 0.91& 3.54 & 9.84 & ~~0.318 & ~~0.445 & ~~\textbf{0.522} & 16.0 \\
    \bottomrule
    \end{tabular}
    \caption{Performance on ML-G2D test set when incorporating different types of training data, including \textbf{ReDial} training data, \textbf{AUGUST} synthesized data, and \textbf{ML}-G2D training set.}
    \label{tab:rec_eval}
    \end{threeparttable}
\end{table*}
\begin{table}[t]
    \centering
    \setlength{\tabcolsep}{2.1mm}
    \setlength{\abovecaptionskip}{2mm}
    \begin{tabular}{cc|ccc}
    \toprule
    % \small \textbf{ReDial} & \small \textbf{AUGUST} & \small \textbf{R@1} & \small \textbf{R@10} & \small \textbf{R@50} \\
    \textbf{ReDial} & \textbf{AUGUST} & \textbf{R@1} & \textbf{R@10} & \textbf{R@50} \\
    \midrule
               &            & 0.0 &  0.0 &  0.0 \\
               & \checkmark & 2.5 & 15.7 & 33.2 \\
    \checkmark &            & 3.9 & 18.3 & 37.8 \\ 
    \checkmark & \checkmark & 3.2 & 17.8 & 36.6 \\
         F     &      P     & \textbf{5.3} & \textbf{25.1} & \textbf{47.1} \\
    \bottomrule
    \end{tabular}
    \caption{ Recommendation accuracy on the ReDial test set when trained on the ReDial and AUGUST data. }
    \label{tab:rec_redial}
\end{table}

\subsubsection{Datasets}
To validate the Data2Text generation quality of AUGUST, we construct graph-dialogue pairs from the ReDial \cite{li2018towards} and WebNLG~\cite{gardent2017webnlg} dataset for training and evaluation. Considering the limitations of existing datasets as stated in Sec.~\ref{intro}, we create a small dataset with more ``real-world'' and reliable recommendations for CR evaluation. We sample 200 pieces of user-item data from MovieLens and hire some annotators to create conversations according to the user preferences for the movies, named ``ML-G2D'' in Tab.~\ref{tab:auto_eval}. We also provide annotators with external knowledge (\eg, movie websites) and ReDial dialogue samples as references to guarantee conversation quality. Among the annotated 200 dialogues, 100 are randomly sampled and used for training in the low-resource scenario, and the other 100 are set as the test set. Note that when testing on WebNLG in Tab.~\ref{tab:auto_eval}, we use WebNLG as the dialogue resource to train the Data2Text generator in AUGUST, and when testing on ReDial and ML-G2D, we both use ReDial as the dialogue resource.
To validate the benefit of synthesized data by our AUGUST, we implement experiments to use our synthesized data as training data for the CR task. Note that to compare the benefit brought by the synthesized data and ReDial data, we randomly sample around 8,000 pieces from the synthesized data for the later training of KGSF, which keeps the same scale as ReDial training data. The synthesized data is denoted as ``AUGUST'' in Tab.~\ref{tab:rec_eval} and \ref{tab:rec_redial}. 
% To evaluate the performance of our methods on Graph2Dialogue and general Data2Text generation tasks, we conduct experiments on three datasets including ReKG2D, MLKG2D, and WebNLG~\cite{gardent2017webnlg}. Each sample consists of an input graph formed by triples and a target text in the form of sentences or dialogues along with a mapping from entity mentions to corresponding KG entities.  The training instances in ReKG2D and MLKG2D are both graph-dialogue pairs reformalized from the ReDial dataset,  while the test instances of ReKG2D and MLKG2D are Graph-Dialogue pairs generated from ReDial and MovieLens, respectively. For evaluation, we asked human annotators to write 100 conversations based on triples sampled from MovieLens as the generation targets. WebNLG is a popular benchmark for a general Data2Text generation task, which contains less input triples and shorter target text compared with ReKG2D and MLKG2D. 

%Table \ref{}shows the statistics for the generated dataset.
\subsubsection{Evaluation Metrics}
To investigate the performance of various methods on the Data2Text generation task, we first conduct evaluations on the quality of \textbf{conversation reconstruction}. We adopt four automatic evaluation metrics widely used in Data2Text generation tasks~\cite{li2021few}: BLEU~\cite{papineni2002bleu} and ROUGE-L~\cite{lin2004rouge}, which computes the overlap ratio of $n$-grams between the reconstructed dialogue and the original one; CIDEr~\cite{vedantam2015cider} that computes the TF-IDF weights for each $n$-gram in synthetic/real dialogues; and Chrf++~\cite{popovic2017chrf++} that computes the average F-score on both character-level and word-level $n$-grams. In addition, we also compute the recall ratio (Recall) of entities to measure how many entities are recovered in the dialogue relative to the graph input. 
For the \textbf{conversation writing} quality, we compute Dist-$n$~\cite{li2015diversity} to show the distinctness of the generated utterances and the perplexity (PPL) proposed in \cite{jelinek1977perplexity} to measure the language fluency. Besides, we also conduct human evaluation to show the generation quality following the previous works in \cite{li2021few, agarwal2021knowledge}, which contains three workers' ratings to 200 randomly sampled dialogues with respect to language naturalness including aspects of fluency, dialogue logic, and informativeness (5 is the full score). 
As for the \textbf{evaluation of CRS} trained on the synthesized data by AUGUST, we follow \cite{li2018towards, chen2019towards, zhou2020improving} to use Recall@$k$ (R@$k$, $k=$ 1, 10, 50) as the recommendation evaluation metric, which indicates whether the predicted top-$k$ items contain the ground truth recommendation provided by human recommenders. The generation quality of CRS is evaluated on Dist-$n$ and PPL as in the Data2Text generation task.
% we conduct human evaluation in addition to automatic evaluation, we randomly sample 200 KG subgraphs along with corresponding generated text from baseline models and our model. Three workers were asked to score the text with respect to two aspects: Factual correctness and Language naturalness to evaluate how well the generated conversation correctly conveys information contained in the KG and how well the generated conversation is grammatically correct and fluent.

%\subsubsection{Models in this study}
%We investigate BART \cite{lewis2020bart} and T5 \cite{raffel2020exploring}, two PLMs based on the Transformer encoder-decoder architecture \cite{vaswani2017attention} for graph-to-conversation generation as previous works \cite{ribeiro2021investigating,li2021few} on graph-to-text generation tasks. 
%To evaluate various models' performance on generating conversation from the given graph, We chose to explore and compare Transformer, Pointer Generator, BART, T5, and our model's performance on data-to-text generation task.

\subsubsection{Implementation Details}
\label{sec:imp}
In the step of Data2Text generation, the graph encoder in AUGUST is implemented as a two-layer R-GCN with an embedding size of 1,024. The PLM encoder for context-aware entity embedding adopts the encoder of a pre-trained BART-large~\cite{lewis2020bart}, which is a transformer-based model with a bidirectional encoder and an autoregressive decoder. The initial weights are provided by Hugging Face\footnote{https://huggingface.co/facebook} and are frozen in training. As for the text decoder, we employ the decoder of a BART-large initialized with pre-trained weights for dialogue generation. The parameters in the R-GCN encoder and BART decoder are optimized using an AdamW~\cite{loshchilov2017decoupled} optimizer with a learning rate of $10^{-5}$. The weight factors, $\lambda_1$ and $\lambda_2$, are set to 0.8 and 0.8, respectively. The whole network is trained on 4$\times$23GB NVIDIA Tesla P40 with a minibatch size of 16. To validate the benefit of synthesized data by AUGUST, we implement a popular CRS, KGSF~\cite{zhou2020improving}, as the baseline, which incorporates two KGs, ConceptNet~\cite{speer2017conceptnet} and DBpedia~\cite{auer2007dbpedia}, to enhance the data representations. Implementation details can be referred to in the released codes by \citeauthor{zhou2020improving}\footnote{https://github.com/Lancelot39/KGSF}. 
%\cite{schlichtkrull2018modeling}
%settings hyper-parameters

\subsection{Experiment Results}
\subsubsection{Data2Text Evaluation}
We give both automatic and human evaluations of the generation quality by AUGUST. For automatic evaluation, we implement AUGUST with BART-large as the PLM, on all three datasets to construct a benchmark for future related works. As shown in Tab.~\ref{tab:auto_eval}, with the same training data, AUGUST performs poorer on ML-G2D than on Re-G2D, which may result from the distribution bias of ReDial data with real-world user preferences as stated in Sec.~\ref{intro}. Besides, the PPL values are low in all settings, so the generation has high confidence, which may result from the consistency of the generation objective between BART pre-training and Data2Text training. Performances on WebNLG are higher than on the other two over all metrics except PPL, because the target text in WebNLG is usually shorter and with richer common entities, and the input has fewer triples, which reduces the generation difficulty. Besides, we also directly compare the quality of the synthesized data by AUGUST and the ReDial data, on ``Distinctness'' and ``Language Naturalness'' in Tab.~\ref{eva:human}. We compute the Dist-2 and Dist-3 scores, and conduct human evaluation on the dialogue logic, fluency, and informativeness, which shows that the synthesized data has a high quality that is close to the ReDial data on both utterance distinctness and language naturalness.
% Considering there is no standard answer for the generation task, automatic evaluation makes limited demonstration for the generation. Thus, we further introduce human evaluation for the generation results by AUGUST and human-labeled dialogues from the ReDial dataset. The human evaluation is conducted on two aspects: (i) Factual Correctness, \eg~the recall of input entities or relations; (ii) Language Naturalness, including expression logic, fluency, grammar accuracy, and informativeness. As presented in Tab.~\ref{eva:human}, the full score for all metrics is 5, and AUGUST implemented with Bart-large performs slightly better than with Bart-base in factual correctness. With respect to the language naturalness, impressively, generations from AUGUST are of comparable quality with samples from the ReDial dataset, which shows the great potential of automatic generation for CR datasets. 

\subsubsection{CR Evaluation}
We evaluate the CR performance of KGSF on the ML-G2D test set, with using different types of data in training. The training data is a combination of external data: ReDial training set and AUGUST synthesized data, and internal data: ML-G2D training set. We set the ratio of the used ML-G2D data to 0\%, 50\%, and 100\% to investigate the performance in low-resource scenarios with different extents. From the results in Tab.~\ref{tab:rec_eval}, it can be seen: (i) KGSF without any external training data (ReDial or AUGUST) performs poor on recommendation; (ii) Using only ReDial as external data can bring benefits to the conversation generation, but leads to only a tiny improvement in recommendation; (iii) Using only AUGUST as external data can bring a significant improvement in recommendation compared with ReDial, especially in a more low-resource scenario; (iv) Using both ReDial and AUGUST as external data cannot bring extra gains on recommendation accuracy but can improve the distinctness in the generated conversations. These results show that: (1) There exists a distribution bias between the recommendations in ReDial data and the user preference in the real world, which results in the unsatisfying recommendation performance of a ReDial-trained CRS; (2) The synthesized data by AUGUST is useful to help a KGSF capture real-world user preferences for conversational recommendations, especially in low-resource scenarios; (3) ReDial and AUGUST data are complementary to provide a more rich corpus for improving the conversation capability of CRS, and adding AUGUST data also leads to a higher recommendation accuracy than using ReDial data only. 

We also evaluate the recommendation performance of KGSF on the ReDial test set when trained with ReDial or/and AUGUST data. As shown in Tab.~\ref{tab:rec_redial}, it can be seen: (i) The recommendation accuracy of KGSF is low without any training data; (ii) Adding synthesized AUGUST data can bring performance gain to get close to but lower than adding real ReDial training data; (iii) Simply adopting joint training with ReDial and AUGUST data can  only obtain similar performance as using ReDial data only; (iv) Using AUGUST data as pre-training and finetuning on ReDial data can bring an extra performance gain. The results of (ii) further prove the benefit of synthesized data by AUGUST and the distribution bias between ReDial recommendations and real-world user preferences. In addition, although simply jointly using both data for training can hardly bring performance gain as in (iii) considering the distribution bias, the synthesized AUGUST data can still help improve the recommendation ability of KGSF when using AUGUST data for pre-training and finetuning on ReDial data. In this way, the AUGUST data provide a better initialization for the optimization of KGSF, and finetuning on ReDial data can guarantee the distribution consistency. It also shows the great potential of AUGUST to serve as a data synthesis approach for a better initialization of parameters in CRS.

\begin{figure}[t]
\centering 
\includegraphics[width=0.478\textwidth]{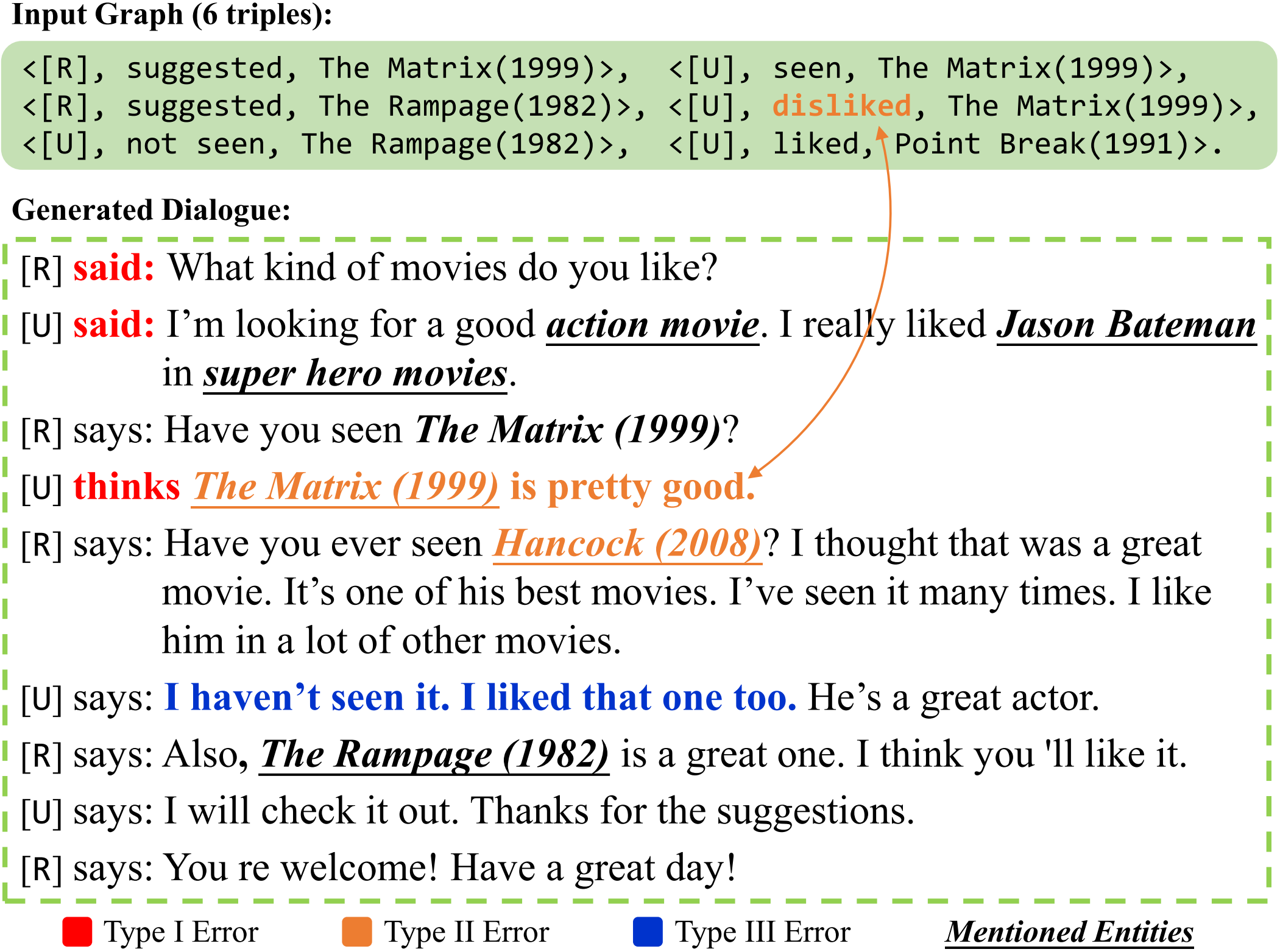}
\caption{Visualization of a generation case by AUGUST for error analysis.} 
\label{Fig.case} 
\end{figure}

% \subsubsection{Ablation Study}
% We conduct ablative experiments to validate the effect of each component, \ie, representation alignment (RA), domain adaptation (DA), and copy mechanism (CP), which correspond to three loss items, \ie, $L_{align}$, $L_{adv}$, and $L_{copy}$, respectively. The basic model is implemented with Bart-large, and all variants are evaluated on the ReKG2D test set over five metrics. As shown in Tab.~\ref{ablation}, the performance drops over almost all metrics without any of the components. Specifically, suppressing the RA component has a relatively small influence on the performance; the DA component mainly contributes to the ROUGE metric, which measures the quality of recovery compared with the reference; and the CP component can significantly affect CIDEr, Chrf, and Recall, which shows the importance of forcing the model to include the input items in generation. 

% \input{body/tab/human_eval}
% \input{body/tab/ablation}

%\subsubsection{Case Study}

\subsubsection{Error Analysis}
\label{error}
We summarize three types of errors that appeared in our generation according to the hierarchy of the dialogue requirement, with one example shown in Fig.~\ref{Fig.case}.
\textbf{Error Type I: Format Errors}, including grammar and spelling mistakes, or the unexpected writing format, \eg, each utterance is expected to start with the identity of ``[U] says:", while it may generate ``[U] thinks".
% the problem that a system generates text which say things are not true, or at least are not in the input data. Specifically, it can be i) a wrong relation between two mentioned entities, (\eg, the truth in the given input triples is that the``Seeker" doesn't like the movie "the Matrix" yet  the "Seeker" shows a fondness for ``the Matrix" in the generated dialogues, since the lack of fine-grained mappings between input triples and the conversation); ii) missing mention of one or several entities from the given input; iii) wrong reference to an entity that is not in the input; iv) wrong description of an entity (\eg, generate wrong genre, director, or synopsis of a movie).
\textbf{Error Type II: Hallucination}, which is a common problem in language generation tasks. It means the network (i) generates contents that conflict with the input data, \eg~producing wrong relations, entities, or sentiments, or (ii) generates extra items beyond the input, which means the output is not a precise description to the input, \eg~``Hancock (2008)'' in Fig.~\ref{Fig.case}. 
\textbf{Error Type III: Incoherent Logic}, which refers to the problem of incoherent  or contradictory logic in the generated dialogue, \eg~the user says (s)he has not seen a movie but liked it.

% Please add the following required packages to your document preamble:
% \usepackage{multirow}
% Please add the following required packages to your document preamble:
% \usepackage{multirow}

\section{Conclusion}
%第二conwe propose a baseline framework on this task with a KG-enhanced PLM, and investigate the effect of each moule. Experiment results shows that our generation is comparable to human labeled conversation with other advantages in ability of scalable, extandable, explainable, and we believe such a task transition brings innovative potential to CRS. 

This paper proposes an automatic generation understudy for conversational recommendation datasets. By casting the dialogue synthesis process as a Data2Text generation task, a baseline framework is constructed to exploit (i) rich accurate user preferences from user-item matrices, (ii) rich external knowledge from external knowledge graphs, and (iii) the conversation ability from the corpus of existing CR datasets. Experiment results show that our generation is comparable to human-labeled conversations and superior in scalability, extensibility, and explainability. More importantly, we empirically show the benefit of our synthesized data in improving a CRS, especially in recommendation accuracy. The proposed approach exhibits great potential for automatic dataset synthesis and is expected to inspire researchers in other fields. 
% We expect to release a conversational recommendation dataset with a grounded user-item graph from the observed real-world users.  We expect future work on: 1) human verifying for guarantee the quality of automatically generated dialogue to finally organized into a ready-to-use dataset or as a supplemental dataset; 2) proposing automatic evaluation metrics for evaluating models' performance on generating conversation from structured data since simply computing the similarity between generation and reference can not match the essential necessity of this task and thus is less significant; 3) exploring more on the structure or representation of the ``dialogue skeleton", an ideal skeleton should be able to both outline our body shape and be flexible enough to fill in different appearances; 4) transferring and expanding this framework to other domain.   
%\subsection{future work}
%\paragraph{Resource}
%\paragraph{Conversation generation}
%\textbf{Manually Verified} 
%\textbf{Reliability} although the conversation can be generated (simulated), the generated questions and user response should be reasonable. 
%\textbf{Toward goal-oriented}
%\textbf{Trade-off between Controllability and Generalizability}
%\paragraph{Evaluation metrics}
%\paragraph{Domain Transfer or Application extension}

\section*{Limitations}
The limitations of this work mainly lie in two aspects: (i) The synthesis quality is determined by the performance of existing Data2Text approaches, while Data2Text generation is still a difficult task that waiting for deeper exploration. The common errors in generation are included in Sec.~\ref{error}. (ii) We adopt a PLM as the decoder in Data2Text generation in order to generate fluent utterances. However, as stated in \cite{ribeiro2021investigating}, PLMs tend to pay more attention to sentence fluency than to the graph structures of inputs, which may cause the loss of some critical information.
% ACL 2023 requires all submissions to have a section titled ``Limitations'', for discussing the limitations of the paper as a complement to the discussion of strengths in the main text. This section should occur after the conclusion, but before the references. It will not count towards the page limit.
% The discussion of limitations is mandatory. Papers without a limitation section will be desk-rejected without review.

% While we are open to different types of limitations, just mentioning that a set of results have been shown for English only probably does not reflect what we expect. 
% Mentioning that the method works mostly for languages with limited morphology, like English, is a much better alternative.
% In addition, limitations such as low scalability to long text, the requirement of large GPU resources, or other things that inspire crucial further investigation are welcome.
\section*{Acknowledgement}
The work was supported in part by NSFC with Grant No. 62293482, the Basic Research Project No. HZQB-KCZYZ-2021067 of Hetao Shenzhen-HK S$\&$T Cooperation Zone,  the National Key R$\&$D Program of China with grant No. 2018YFB1800800, the Shenzhen Outstanding Talents Training Fund 202002, the Guangdong Research Projects No. 2017ZT07X152 and No. 2019CX01X104, the Guangdong Provincial Key Laboratory of Future Networks of Intelligence (Grant No. 2022B1212010001), the Shenzhen Key Laboratory of Big Data and Artificial Intelligence (Grant No. ZDSYS201707251409055), and the National Key R$\&$D Program of China under Grant No. 2020AAA0108600.

\bibliography{anthology,custom}
\bibliographystyle{acl_natbib}

% \appendix

% \section{Example Appendix}
% \label{sec:appendix}

% This is a section in the appendix.

\end{document}